# LIFE - an energy efficient advanced continual learning agentic AI framework for frontier systems

Anne Lee
*AI Research Lab*
*Nokia Bell Labs*
Naperville, USA
anne.lee@nokia-bell-labs.com

Gurudutt Hosangadi
*AI Research Lab*
*Nokia Bell Labs*
Murray Hill, USA
gurudutt.hosangadi@nokia-bell-labs.com

*Abstract*—The rapid advancement of AI has changed the character of HPC usage – dimensioning, provisioning, and execution. Not only has energy demand been amplified, but existing rudimentary continual learning capabilities limit AI's ability to effectively manage HPCs.

This paper reviews emerging directions beyond monolithic transformers, emphasizing agentic AI and brain-inspired architectures as complementary paths toward sustainable, adaptive systems. We propose LIFE—a reasoning and Learning framework that is Incremental, Flexible, and Energy efficient—implemented as an agent-centric system rather than a single monolithic model. LIFE uniquely combines four components to realize self-evolving network management and operations in HPCs. The components are an orchestrator, Agentic Context Engineering, a novel memory system, and information lattice learning. LIFE can also generalize to enable a variety of orthogonal use cases.

We ground LIFE in a specific closed-loop HPC operations example - detecting and mitigating latency spikes experienced by critical microservices running on a Kubernetes-like cluster.

## I. INTRODUCTION

The rapid advancements in artificial intelligence (AI) have brought forth remarkable capabilities. However, they have also exposed critical challenges to successfully achieve our goals for autonomous self-evolving systems such as future HPC Data Centers — foremost amongst them are the energy demands of AI models especially as they've scaled upwards [1,2]. It is not just large language models (LLMs) that have grown to trillions of parameters. The number of small/medium language models (SLMs and MLMs) is growing rapidly as well with an aggregate energy consumption that could rival that of LLMs. As intelligent agents powered by these models begin to play a key role in building, deploying, and optimizing HPC applications, their energy efficiency will be a key factor in AI driven sustainable HPC.

Another deficiency is the immaturity of continual learning or the rudimentary ability of AI models to adapt in non-stationary environments. It is especially critical for HPC platforms to be able to deal with extreme complexity, optimize resource utilization in real-time, and shift from reactive, human-intensive management to autonomous, proactive intelligent operations. These two challenges, energy efficiency and continual learning are closely coupled. Models that can learn incrementally, adapt to new conditions, and selectively retain knowledge can potentially reduce the need for expensive retraining significantly. Figure 1 illustrates the close linkage between continual learning and energy efficiency and how they impact agent and model design given data and physical constraints. Learning efficiency—defined as the ability to acquire and integrate new knowledge using minimal data, computation, and energy—is therefore a critical design goal for next-generation AI systems.

In the context of future data center networks, these challenges become central rather than peripheral. AI-nativeness and sustainability are foundational pillars of future network infrastructures. Continual learning is key to reaching full system autonomy. This paper explores emerging AI paradigms around two promising paths: agentic AI and brain inspired AI. Specifically, to address the outlined challenges, we propose an agent-centric framework that combines energy-efficient learning principles, externalized memory, and adaptive AI models to support advanced continual learning.

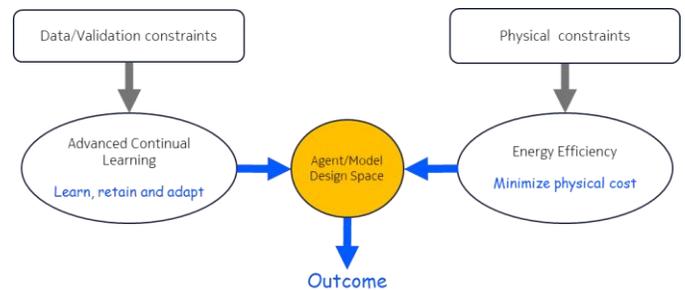

*Figure 1 Continual learning and energy efficiency will drive future Agent designs*

It is NOT the suggestion or expectation of this paper that brain-inspired AI approaches will be the only solution for minimizing energy usage or making continual learning more effective. After all, there are already known approaches being investigated and/or currently used that could result in lower energy consumption and some level of continual learning. But, they are not, by themselves, sufficient. We will, however, include a summary of these methods along with our proposal later in the paper after discussing our main goals that drive the rest of the paper.

## II. GOALS AND TENETS

### A. Goals

Our proposal focuses on a foundational learning system that supports two simultaneous goals – energy efficiency and continual efficient learning with adaptive flexibility:

1. Energy efficiency:
   - Dramatically reduce energy consumption during both training and inference which is also characteristic of the human brain.
   - This will be an enabler of sustainable AI driven HPC operations.
2. Continual learning with adaptive flexibility which can be broken into sub-goals:
   - Stability: the ability to learn incrementally over time while retaining previously acquired knowledge.
   - Plasticity: adapt rapidly to short-term and long-term changes without catastrophic forgetting.
   - Ensure an optimal trade-off between stability, plasticity and generalizability to intra-task and inter-task distribution differences [4].

A distinction between continual learning and reinforcement learning should be noted here. The latter is primarily focused on finding the optimal policy for a fixed or specific task while the former focuses on lifelong learning without forgetting previously learned knowledge. The two can be combined that could lead to continual reinforcement learning (CRL) [5] which underpins Richard Sutton's drive to super-intelligence. He has proposed a new architecture called "Options and Knowledge" (OaK) for building agents that can reach superintelligence by [continually] learning from experience.[6] Our proposal of focusing on advanced continual learning combined with energy efficiency has humbler ambitions than superintelligence.

Furthermore, we will concentrate on achieving these goals in the context of tasks for systems such as data centers at two levels. First, AI models for specific tasks, and second, the agents that can operate in an autonomous way using these AI models.

### B. Tenets

Achieving continual learning with energy efficiency requires adherence to a set of guiding principles that apply across models and agents. By adhering to these principles, we can create AI systems that are not only powerful and effective but are also capable of evolving alongside the dynamic environments they are designed to navigate.

*1) Energy efficiency*

One path in the pursuit of energy efficiency is for models to increasingly emulate the brain's remarkable ability to optimize energy use. This drives the guiding principles for energy efficiency in AI models which include:

- Sparsity, where only a small fraction of neurons is active during any given task.
- Event-driven neuron activation, where neurons are triggered only by specific events rather than continuous activation.
- Localized learning, learning functions execute directly where the data resides.
- Minimized data movement, reducing energy-intensive transfers between memory and compute using techniques such as in-memory computing.
- Architectural co-design such as using neuromorphic hardware, which can play a pivotal role in overcoming the energy bottlenecks of traditional computing architectures.

The above principles are inspired in part by biological systems but are increasingly reflected in emerging AI architectures and hardware platforms.

*2) Continual learning with adaptive flexibility*

Continual learning systems must balance stability and plasticity across multiple time scales:

- Short term learning, enabling adaptation to temporary events with conditional forgetting of short duration stimuli and on-the-go transfer learning
- Long term learning, allowing gradual adaptation to persistent knowledge or experience without overwriting of previous learned knowledge except in specific situations (i.e. conditional forgetting is okay).

AI systems, much like the human brain, must balance their limited capacity with the need to learn new information, whether temporarily or permanently. This requires a dual focus on short-term and long-term learning, each with its own unique challenges and solutions.

In short-term learning, models must exhibit flexibility to handle temporary events. For instance, a Site Reliability Engineering (SRE) model tuned to diagnose Kubernetes microservice latency may need to temporarily treat a burst of DNS errors or a short-lived Top of Rack (ToR) packet-loss event as the primary signature, without permanently shifting its baseline understanding of the service. Conditional forgetting supports this by letting the model discard these short-duration correlations once the event passes while preserving foundational knowledge about normal dependency behavior. Additionally, on-the-go transfer learning allows insights from one incident (e.g., ingress throttling) to rapidly bootstrap troubleshooting for a related service or cluster without requiring long-term retention of all incident-specific details.

Long-term learning, on the other hand, requires models to adapt to non-stationary data streams while avoiding catastrophic forgetting. Models must continuously learn from evolving telemetry and systems without retraining on the entire history. For example, an incident classifier initially trained on a particular cluster and software stack (specific Kubernetes

Identify applicable funding agency here. If none, delete this text box.

version, service mesh, and hardware generation) can be incrementally updated as new versions, workloads, and failure modes appear—such as new sidecar behaviors, GPU node types, or storage backends—while retaining prior diagnostic skill. Conditional forgetting also plays a critical role given limited capacity: the system should discard outdated knowledge (e.g., alerts tied to decommissioned hardware or retired configurations) to make room for current operational patterns.

Note that while we specifically call out AI models in this discussion, these points are equally relevant to AI agents. In fact, solutions to current limitations may lie in new components or architectures of an AI agent rather than rely solely on the AI model.

### III. STATE OF THE BUSINESS AND LIMITATIONS

The focus here is specifically around the goals mentioned in the previous section, namely energy efficiency and continual learning with adaptive flexibility. Since transformers are the predominant SOTA AI models, we will begin by looking at methods developed for them that are proposed to support our stated goals. This will be followed by investigations into new brain-inspired AI models along with combining the models with agentic approaches.

#### A. Evolution of transformers

With respect to energy efficiency, various methods are available including using fewer data samples for training (e.g. few-shot learning), quantization, task-specific small AI models, pruning and knowledge distillation to shrink the model. There are also chips being developed specifically to run these AI models more efficiently. With respect to continual learning, the top and emerging approaches currently are listed in Table 1.

*1) Model-level vs Agent-level Continual Learning*

Some approaches are designed to operate inside the AI model (e.g., regularization, replay, architectural expansion). Others can operate at the AI agent level, enabling continual learning through external memory, context management, or tool orchestration without modifying the model itself.

*2) Selecting a Continual Learning Method*

Choosing an approach depends on the requirements of the use case, including:

- Real-time learning requirements
- Energy or compute constraints
- Short-term vs long-term adaptation
- Deployment location (inside the model vs within an AI agent)
- Model compatibility (transformers, classical ML, or other architectures)

Table 1 also describes which requirement, each listed method supports:

*Table 1 Sample Continual Learning Methods*

| Category | Concept | Method | Real-time | Short-term | Long-term | Energy reduction | Model or Agent |
|---|---|---|---|---|---|---|---|
| Regularization & stability | Prevent catastrophic forgetting by constraining parameter updates. | Elastic Weight Consolidation (EWC), Synaptic Intelligence (SI) | P | P | YES | YES | Model |
| | | Learning without Forgetting (LwF) | P | YES | P | P | Model |
| Replay & memory | Reuse past information through stored or generated data. | Experience replay / rehearsal | P | YES | P | P | Model/ Agent |
| | | Generative replay | N | P | YES | N | Model |
| | | Gradient-based (GEM, A-GEM) | P | YES | YES | P | Model |
| | | **Agentic Context Engineering (ACE)** | **YES** | **YES** | **YES** | **YES** | **Agent** |
| Parameter isolation / arch. | Allocate or reuse dedicated model capacity for new tasks. | Progressive neural networks | N | P | YES | N | Model |
| | | Dynamic architecture expansion (DEN, etc.) | P | P | YES | N | Model |
| | | PackNet, Piggyback, Supermasks | P | YES | P | YES | Model |
| | | PEFT – Sparse Memory Fine-Tuning (SMFT), LoRA & prompt adapters | P | YES | YES | YES | Model/ Agent-selected |
| | | Orthogonal gradient methods (OGD, etc.) | P | YES | YES | P | Model |
| Meta & orthogonal gradients | Adapts to new tasks using examples in the prompt without updating model parameters, enabling short-term adaptation. | Meta-Continual Learning (e.g., Nested Learning) | P | YES | YES | Y (amortized) | Model |

(Legend: **Y** = Yes/natural fit, **P** = Possible/partial/with caveats, **N** = Generally no / poor fit)

*B. Brain-inspired AI model architectures*

Various AI architectures have been called brain-inspired including neural networks. There has been criticism, however, that some of these architectures don't mimic the brain as closely as is implied.

In this section, we are focused on AI model architectures that truly mimic specific brain mechanisms. These are listed in Table 2 along with open source implementations.

*Table 2 Brain-inspired Architecture Approaches*

| Architecture /Code | Description | Energy Efficiency | Continual Learning | Limitations |
|---|---|---|---|---|
| **Spiking Neural Networks (SNN) [11] / [21-24]** | Neurons communicate via discrete "spikes" instead of continuous activations. | Highly energy-efficient, especially if run on neuromorphic hardware | Supports adaptive learning, mitigating catastrophic forgetting. [14] | Accuracy issues with fewer neurons, training challenges due to non-differentiable nature of spikes.[15] |
| **Liquid Neural Networks (LNN) or Liquid Time Constant networks [12] / [19] / [25]** | Fixed recurrent networks (the "liquid") transform signals into a high-dimensional, dynamic representation | More energy efficient than traditional networks due to fewer richer neurons | Strong adaptive learning and robustness compared to traditional networks. | Vanishing gradients; slow training time; scaling issues with large datasets [16] |
| **Hierarchical Reasoning Models (HRMs) [13] / [26]** | Break down complex problems into simpler, interconnected sub-problems. | Avoids the energy-intensive "chain-of-thought" (CoT) token generation used by LLMs. | Dependent on model/algorithms used at each level. Streamlined variants like TRM reduce the system size dramatically | Immature technology; unclear advantages over transformer or diffusion models [17] |

Researchers are also working to streamline architectural approaches, such as the Tiny Recursive Model (TRM) [18], to reduce model size while maintaining high accuracy. Another emerging approach which we mentioned earlier is Rich Sutton's OaK architecture, a model-based reinforcement learning framework aimed at achieving superintelligence through lifelong, continual learning from experience. Unlike traditional methods, OaK avoids backpropagation and static datasets, relying instead on agents that learn through direct interaction with their environment, starting with no pre-programmed knowledge. OaK supports the Alberta Plan, a 12-step initiative to transition AI from "frozen" static models to adaptive, real-time learning agents. Currently, OaK lacks a complete implementation, with foundational algorithms and components scattered across various repositories. During Sutton's keynote at NeurIPS 2025, he invited the AI community to contribute to OaK's development

*C. Combining evolving continual learning paradigms with brain-inspired AI models – beyond transformers*

All of the continual learning methods listed above are not inherently transformer-based and can be used in or with spiking neural networks, liquid neural networks, and hierarchical reasoning models as well as other non-transformer models. There is just one caveat. PEFT (SMFT, LoRA & prompt-based methods) were developed for large dense neural nets (esp. transformers); they are usable only with compatible parameterizations (e.g., learnable linear layers, attention-like blocks) inside the non-transformer model

## IV. LIFE – OUR CONTINUAL LEARNING FRAMEWORK

The following describes our vision and approach to achieve the goal of advanced energy efficient continual learning for AI systems.

*A. Our vision*

Our vision is to develop a human brain inspired learning and reasoning system called **LIFE** implying a **L**earning system or model that is:

1. **I**ncremental (or continual), making adaptation long-term sustainable.
2. **F**lexible (or adaptive), adjusting instantly to new events making continual learning situationally efficient.
3. **E**nergy efficient, making both continual and adaptive learning practically deployable in constrained environments.

Rather than attempting to build a single monolithic model, LIFE is conceived as an agent-centric framework that integrates AI models, externalized memory, and orchestration logic. This approach aligns naturally with the vision of fully autonomous networks and data centers composed of interacting AI agents.

*B. Our approach*

Earlier in this paper, we explored advancements in learning architectures and platforms that aim to achieve continual learning with adaptive flexibility and energy efficiency. While these advancements represent significant progress, they each come with inherent limitations as shown in Table 1 and Table 2.

The paper highlights the challenge of balancing learning efficiency and energy efficiency in AI, as advanced models often require significantly more energy. Incremental solutions like optimized batch sizes and efficient hardware help but are insufficient for growing AI demands. To address this challenge, we believe the most effective path forward lies in combining multiple approaches tailored to specific tasks.

Success in AI should be measured using a general metric encompassing learning and energy efficiency across tasks, rather than narrow metrics like accuracy and energy use in specific tasks (e.g., object detection [20]). Additional research is needed to develop benchmarks aligned with the proposal's main research directions.

*C. Our proposed architecture*

The LIFE framework distinguishes between two complementary innovation paths:

- Inside the model, using energy-efficient and continually adaptable reasoning methods.
- Outside the model, through agent-level mechanisms that manage context, memory, and learning without requiring direct access to model internals.

The latter path provides a separation that enables LIFE agents to operate with open-source models, proprietary models, or third-party APIs. This approach aligns well with an autonomous AI-native system which will consist of a collection of AI models and AI agents some of which will interact peer-to-peer with each other and in other cases will be hierarchically connected to each other, essentially creating an "internet-of-agents" that operates the system. Each system task will be implemented with an AI model or AI agent that is customized for its function.

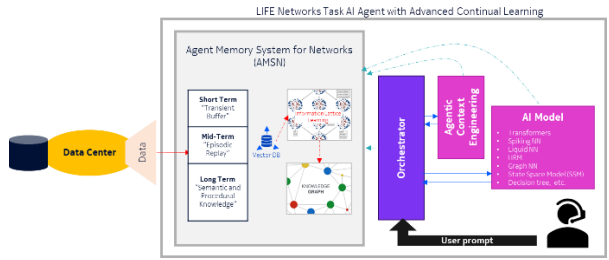

*Figure 2 LIFE AI Agent framework*

Figure 2 illustrates our proposed integrated LIFE AI agent solution using a neuro-symbolic approach.

*1) End-to-end workflow*

The following is the end-to-end workflow through our Figure 2 LIFE AI Agent framework.

1. **User → Orchestrator:** User prompt is received by the orchestrator.
2. **Orchestrator ↔ ACE:** Orchestrator queries/updates ACE to prepare context (e.g., history, tools, policies) for the model, possibly through multiple interactions.
3. **Orchestrator → AI Model:** Orchestrator sends the user prompt, curated context, and instructions to the AI model.
4. **AI Model → Orchestrator → User/Tools:** Model output returns to the orchestrator, which decides to respond, call tools/agents, or loop back through ACE and the model

*2) Agent Memory System for Networks (AMSN)*

In order to meet the requirements for continual learning that we have described earlier in the paper, a robust memory system is needed. At the core of LIFE is the Agent Memory System for Networks (ASMN), which supports multiple forms of memory:

- Short-term (transient) memory, providing the immediate working context for reasoning.
- Mid-term (episodic) memory, capturing time-stamped experiences and outcomes. This can be used for replay in fine-tuning/retraining, reinforcement learning, and patterns and rules discovery
- Long-term (semantic) memory, representing structured, persistent knowledge such as network topology and policies.
- Procedural memory, storing validated actions and workflows, e.g. from previous successful troubleshooting scenarios.

These memory layers are implemented using a combination of

- Context buffers, containing the instructions (prompts), previous user messages, previous responses, retrieved documents, etc
- Vector databases, specialized databases designed to store, index, and search high-dimensional numerical vectors
- Knowledge graphs, storing structured, interconnected facts about things (entities) and how they relate to each other, usually in a graph form of nodes and edges

Indeed, to design an effective autonomous network with advanced continual learning, it is important to have a sufficiently expressive and extensible ontology that may work in conjunction with a digital twin which is a living instance of that ontology. These would be stored in the knowledge graph.

In addition to these memory components, we introduce Information Lattice Learning (ILL) into LIFE. ILL is a framework for extracting structured, verifiable rules from heterogeneous data sources by organizing concepts into partial-order lattices based on shared attributes, using Formal Concept Analysis as its mathematical foundation [21]. In our LIFE pipelines, ILL distills high-dimensional data, e.g. episodic memory, from vector databases into validated rules, which are filtered based on support (frequency), confidence (reliability), and ontological consistency, and injects them into the knowledge graph as trustworthy semantic assertions. This enables continuous, automated ontology evolution as the network encounters new concepts, relationships, and operational patterns.

*3) Orchestrator*

The orchestrator plays a critical role in managing and optimizing the operations of an agent. It oversees the agent's operational cycle, ensuring smooth transitions through the phases of observing, reasoning, and acting. Additionally, it manages the agent's state machine, handling transitions, loops, and other dynamic processes. The orchestrator is responsible for breaking down complex requests into smaller, manageable subtasks, enabling efficient execution.

A key capability of the orchestrator is deciding which tools, APIs, or capabilities to invoke, such as utilizing Model Context Protocol (MCP) for tool activation. It is also equipped to detect errors and initiate recovery actions to maintain system stability. Resource management is another vital responsibility, as the orchestrator handles API limits, token budgets, and the context window to optimize performance.

Furthermore, the orchestrator executes predefined rules or learned policies, making informed decisions about when to stop, escalate, or prioritize competing objectives. It continuously monitors and logs actions and outcomes to ensure accountability and traceability. Lastly, the orchestrator facilitates communication and collaboration between AI agents using Agent-to-Agent (A2A) interaction, enhancing overall functionality and problem-solving capabilities.

4) *Agentic Context Engineering (ACE)*

Agentic Context Engineering (ACE) is integrated into LIFE to enhance continual learning. It complements the orchestrator. ACE dynamically manages context and optimizes agent interactions to adapt to evolving tasks and environments. ACE gives the orchestrator the right, structured context to make good decisions, while the orchestrator decides what to do next and which tools/agents to call based on that context.

5) *AI model*

AI models within LIFE are task-specific and may include small transformers, graph neural networks, liquid neural networks, decision trees, or spiking models. These may be reasoning models, reasoning-adjacent models, pattern recognition models, signal processing models, etc. Model selection is guided by task requirements such as latency, energy budget, interpretability, and adaptability rather than a one-size-fits-all approach. Table 3Table 1 lists some of the continual strategies to minimize retraining.

*Table 3 Continual Learning Strategies for AI Models*

| Strategy | What it does | Catastrophic forgetting Risk |
|---|---|---|
| Fine-tuning only new layers | Freeze backbone, retrain classifier head | Low — backbone preserved |
| Elastic Weight Consolidation (EWC) | Penalizes changes to weights important for old tasks | Moderate mitigation |
| Experience Replay / Rehearsal | Mix old training samples in with new data | Very effective if you retained old data |
| LoRA / Adapter Layers | Add lightweight adapter modules for new classes, freeze base weights | Very low — old weights untouched |
| Progressive Neural Networks | Add new columns for new tasks, lateral connections to old | No forgetting, but model grows |

6) *Putting it all together – the memory flow*

The orchestrator, agentic context engineering (ACE), and AI model in the LIFE agent may all query the vector DB, knowledge graph or the information lattice. One data workflow through all the memory building blocks is illustrated in Figure 3.

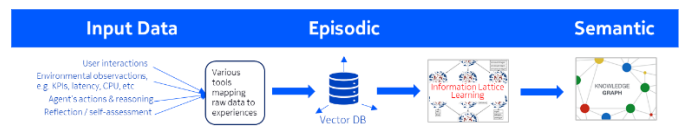

*Figure 3 LIFE AI agent memory pipeline*

D. *HPC data center use case*

LIFE is a framework for an energy-efficient advanced continual learning AI agent framework. This framework can be used to implement AI-native system tasks. We discuss here one possible use case.

1) *High level use guidance*

a) *Data flow in memory component*

Below are examples of how the memory sub-components might be used:
- Vector DB (episodic memory): Retrieve similar cooling system failures during peak load
- Knowledge Graph (semantic memory): Map the fastest path from a virtual machine to storage.
- Information Lattice (pattern/partial-order lattices): Outline energy-saving strategies in cooling systems.

b) *AI task agents*

There will be many different AI agents created using this framework to execute different tasks. Below are some examples:

| HPC data center task agents |
|---|
| Fault Diagnosis Agent |
| Capacity Planning Agent |
| Security Anomaly Agent |
| Resource Usage Optimization Agent |
| Troubleshooting Agent |

*Figure 4 Example LIFE AI task agents for HPC Data Centers*

*c) AI model selection*

The components in the framework for each specific task would be tailored to the task itself. This includes the AI model. Below are guidelines for selecting an AI model depending on the task of the LIFE agent in the context of networks and data centers.

*Table 4 AI model selection*

| AI Model | Use Cases | Benefits |
|---|---|---|
| **Graph neural networks (GNNs)** | • Root Cause Analysis (RCA) for Kubernetes microservice latency spikes using the service dependency graph.<br>• Blast-radius prediction for rollouts/config changes (canary, Horizontal Pod Autoscaler (HPA), mesh, network policy).<br>• Infer contention/noisy neighbors by linking pod placement to node and I/O signals. | Models service/infrastructure dependencies to speed RCA, estimate propagation, and prioritize fixes. |
| **Small, specialized transformers** | • Interpreting tickets, change requests, and logs<br>• Generating explanations, actions/plans | Distilled, domain-tuned models, not giant generic LLMs, to reduce energy and hallucinations. |
| **Liquid neural networks (LNNs)** | • Online anomaly detection on KPI streams.<br>• Predictive maintenance of equipment. | Efficient for non-stationary time series, adapts to continual learning constraints. |
| **Decision trees** | • Simple operational decisions (e.g., threshold-like policies, escalation rules).<br>• Feature-based classification (e.g., "is this alarm combination typical or abnormal?"). | Fast, low-power, and interpretable; ideal for cascades where 80% of events are "easy." |
| **Spiking neural networks (SNNs)** | • Event-driven anomaly detection on high-rate data-center signals (e.g., ToR switch telemetry, Network Interface Card (NIC) drops, power/cooling sensor spikes) to trigger the SRE workflow. | Ultra-low-power, event-driven detection for dense sensor/telemetry streams—ideal as a front-end trigger before heavier RCA models run. |
| **Hierarchical reasoning models (HRMs)** | • Multi-level agent hierarchy<br>• Multi-time-scale memory and learning (short vs long horizon policies). | Supports scaling and stable continual learning. |
| **State Space Models (SSMs) (e.g., Mamba)** | • Long sequence modeling for logs, traces, KPI streams, and time-ordered events.<br>• Streaming anomaly detection / forecasting when long context matters (e.g., diurnal/weekly patterns).<br>• Efficient long-context text understanding for ops workflows (tickets/runbooks) in smaller models. | Linear-time inference enables long-context processing with lower latency and memory than attention, improving efficiency for long event streams. |

*2) End-to-end HPC data center example*

An HPC cluster experiences intermittent latency spikes for a critical microservice running on a Kubernetes cluster. The following shows how our LIFE framework can help diagnose, remediate, and learn from the incident.

*a) Data flow through ASMN*

1. Data Input
    a. Telemetry & Metrics: CPU, memory, disk I/O, network latency, pod restarts, etc.
    b. Events & Logs: Syslogs, Kubernetes events, change records.
    c. Tickets & Alerts: Network Operations Center (NOC) alerts, IT Service Management (ITSM) tickets, SRE discussions.
    d. Processing: Normalize data into a unified schema and classify (performance, capacity, configuration, security).
2. Short-Term Memory (Buffers):
    a. Holds the current query, recent telemetry, retrieved episodes (Vector DB), topology, and policies (KG).
    b. Dynamically updates to support reasoning and action selection.
3. Mid-Term Memory (Vector DB):
    a. Stores past incidents with conditions, actions, and outcomes.
    b. Enables similarity search for relevant historical examples.
4. Long-Term Memory (Knowledge Graph):
    a. Encodes structured knowledge of topology, dependencies, policies, and fault patterns.
    b. Provides context for root cause analysis and impacted components.
5. Procedural Memory:
    a. Stores tested workflows and runbooks for remediation.
    b. Suggests ranked actions based on historical success and compliance.
6. Learning (Information Lattice):

a. Distills episodic data into semantic rules using lattice theory.
  b. Updates the KG with formalized, query-able rules for future reasoning.

*b) Unified workflow*

- Detection: Metrics and alerts identify latency spikes, initiating short-term memory context.
- Enrichment: Episodic memory retrieves similar past incidents; semantic memory provides topology, dependencies, and policies.
- Diagnosis: The agent hypothesizes root causes and identifies impacted services/Service Level Objectives (SLOs).
- Action Selection: Procedural memory suggests workflows, ranked by past success and policy constraints.
- Execution: Procedures are executed, and telemetry monitors SLO restoration.
- Logging: Incident details are stored in episodic memory for future reference.
- Learning: Episodic data is mined for patterns, generating rules to update the Knowledge Graph. This enables self-evolution of the agent as well as forgetting of outdated information
- Improvement: Future incidents benefit from enriched memories, refined workflows, and faster reasoning.

## Conclusion

Practical autonomy in HPC data centers and other AI-native systems is increasingly constrained by two coupled bottlenecks: the energy cost of scaling AI and the limited maturity of continual learning methods that can reduce retraining and ultimately enable self-evolving autonomy. To address these constraints, we introduced LIFE—a Learning framework that is Incremental, Flexible, and Energy-efficient implemented as an agent-centric system rather than a monolithic model. LIFE combines an orchestrator for observe–reason–act loops and budget-aware tool use, Agentic Context Engineering (ACE) to assemble structured context under window limits, and an Agent Memory System for Networks (AMSN) spanning short-term buffers, episodic vector stores, semantic knowledge graphs, and procedural runbooks. We further proposed Information Lattice Learning (ILL) as a neuro-symbolic mechanism to distill episodic experience into validated, ontology-consistent rules that can safely evolve the knowledge graph over time, enabling explainability and long-horizon retention.

Beyond the specific data-center workflow example, LIFE is intended as a general framework for building sustainable frontier systems: push adaptability into agent-level memory, context, and orchestration; reserve heavyweight model updates for cases where they are truly justified; and treat energy, compute, and data movement as first-class constraints. Key next steps include establishing benchmarks that jointly evaluate learning efficiency and energy efficiency across tasks, quantifying end-to-end cost trade-offs among memory tiers and model choices, and validating ILL-driven knowledge updates with rigorous safety checks and human-in-the-loop governance. We believe this agent-centric, neuro-symbolic direction provides a practical path toward self-operating infrastructure that can continually improve while remaining interpretable, auditable, and energy-aware.

## Acknowledgment

The authors would like to thank Sean Kennedy and Matthew Andrews for their review and feedback. We also acknowledge the use of AI, specifically Gemini, Claude, and GPT, for brainstorming, checking facts, help in creating Table 1 and Table 4, help with the conclusion and improving writing style.

Finally, we especially thank Lav Varshney for introducing us to his information lattice learning framework.

## References

[1] S. Mehta, "How Much Energy Do LLMs Consume? Unveiling the Power Behind AI," adasci.org, Jul. 03, 2024. DOI: N/A (web article; no DOI assigned). Available: https://adasci.org/how-much-energy-do-llms-consume-unveiling-the-power-behind-ai

[2] Alex de Vries, "The Growing Energy Footprint of Artificial Intelligence," ScienceDirect, Oct. 18, 2023. Available: https://doi.org/10.1016/j.joule.2023.09.004

[3] Liyuan Wang et.al, "A Comprehensive Survey of Continual Learning: Theory, Method and Application," arXiv:2302.00487v3, Feb. 06, 2024. Available: https://doi.org/10.48550/arXiv.2302.00487

[4] David Abel et.al, "A Definition of Continual Reinforcement Learning," arXiv:2307.11046v2, Dec. 01, 2023. Available: https://doi.org/10.48550/arXiv.2307.11046

[5] Matthias Bastian, "Richard Sutton says the AI industry has 'lost its way' by ignoring core principles of intelligence," the decoder, Aug. 20, 2025. DOI: N/A (news article; no DOI assigned). Available: https://the-decoder.com/richard-sutton-says-the-ai-industry-has-lost-its-way-by-ignoring-core-principles-of-intelligence/

[6] Ben Dickson, "ACE prevents context collapse with 'evolving playbooks' for self-improving AI agents," Venture Beat, Oct. 16, 2025. DOI: N/A (news article; no DOI assigned). Available: https://venturebeat.com/ai/ace-prevents-context-collapse-with-evolving-playbooks-for-self-improving-ai

[7] Ali Behrouz, Student Researcher, and Vahab Mirrokni, VP and Google Fellow, "Introducing Nested Learning: A new ML paradigm for continual learning," Google Research, Nov. 7, 2025. DOI: N/A (blog post; no DOI assigned). Available: Introducing Nested Learning: A new ML paradigm for continual learning

[8] Mitansh Gor, "My reflections on a model that learns how to learn," DEV, Nov. 16, 2025. DOI: N/A (blog post; no DOI assigned). Available: https://dev.to/mitanshgor/nested-learning-my-reflections-on-a-model-that-learns-how-to-learn-14b5

[9] Ali Behrouz et al, "Nested Learning: The Illusion of Deep Learning Architecture," Google Research, 2025. DOI: 10.48550/arXiv.2512.24695. Available: https://abehrouz.github.io/files/NL.pdf

[10] Amirhossein Tavanaei et.al., "Deep Learning in Spiking Neural Networks," arXiv:1804.08150v4, Jan. 20, 2019. Available: https://doi.org/10.48550/arXiv.1804.08150

[11] Ramin Hasini et.al., "Liquid Time-constant Networks," arXiv:2006.04439v4, Dec. 14, 2020. DOI: 10.48550/arXiv.2006.04439. Available: https://arxiv.org/abs/2006.04439

[12] Guan Wang et.al., "Hierarchical Reasoning Model," arXiv:2506.21734v1, June 26, 2025. Available: https://doi.org/10.48550/arXiv.2506.21734

[13] Bing Han et.al., "Enhancing Efficient Continual Learning with Dynamic Structure Development of Spiking Neural Networks," arXiv:2308.04749v1, Aug. 09, 2023. Available: https://doi.org/10.48550/arXiv.2308.04749

[14] Igor Semenov, Dmitry Nikitin, "Advantages and disadvantages of Spiking Neural Networks compared to Artificial Neural Networks," IEEE, Nov. 27, 2023. Available: https://doi.org/10.1109/ICP60417.2023.10397433

[15] Shilong Zong et.al., "Accuracy, Memory Efficiency and Generalization: A comparative study on LNNs and RNNS," arXiv:2510.07578v1, Oct. 08, 2025. Available: https://doi.org/10.48550/arXiv.2510.07578

[16] Renee Ge et.al., "Hierarchical reasoning models: perspectives and misconceptions," arXiv:2510.00355v2, Oct. 07, 2025. Available: https://doi.org/10.48550/arXiv.2510.00355

[17] Alexia Jolicoeur-Martineau, "Less is more: Recursive Reasoning with Tiny Networks," arXiv:2510.04871v1, Oct. 06, 2025. Available: https://doi.org/10.48550/arXiv.2510.04871

[18] Yinlena Xu et.al., "Energy Efficiency of Training Neural Network Architectures: An Empirical Study," arXiv:2302.00967v1, Feb. 02, 2023. Available: https://doi.org/10.48550/arXiv.2302.00967

[19] Ramin Hasini et.al., "Liquid Time-constant Networks," arXiv:2006.04439v4, Dec. 14, 2020. Available: https://doi.org/10.48550/arXiv.2006.04439

[20] Bowen Zhang and Geoffrey Ye Li, "White-Box 3D-OMP-Transformer for ISAC," arXiv:2407.02251v1, Jul. 02, 2024. Available: https://doi.org/10.48550/arXiv.2407.02251

[21] SpikeYolo https://github.com/BICLab/SpikeYOLO DOI: 10.48550/arXiv.2407.20708

[22] SpikingBrain https://github.com/BICLab/SpikingBrain-7B DOI: 10.48550/arXiv.2509.05276

[23] Thousand brains project https://thousandbrains.org/ DOI: N/A (project website; no DOI assigned)

[24] Baby Dragon Hatchling GitHub - pathwaycom/bdh: Baby Dragon Hatchling (BDH) – Architecture and Code DOI: 10.48550/arXiv.2509.26507

[25] Liquid AI https://www.liquid.ai/research/liquid-neural-networks-research DOI: N/A (research webpage; no single DOI assigned)

[26] Sapient Intelligence https://www.sapient.inc/ DOI: 10.48550/arXiv.2506.21734